\documentclass[10pt,twocolumn,letterpaper]{article}

\usepackage{wacv}

\usepackage[utf8]{inputenc} 
\usepackage[T1]{fontenc}    
\usepackage{url}            
\usepackage{booktabs}       
\usepackage{amsfonts}       
\usepackage{nicefrac}       
\usepackage{microtype}      
\usepackage{xcolor}         

\usepackage{chngpage}
\usepackage{multirow}
\usepackage{subcaption}
\usepackage{graphicx}
\usepackage{algpseudocode}
\usepackage{algorithm}          
\usepackage{amsmath,amssymb}    
\usepackage{hyphenat}
\usepackage{color}
\usepackage{mwe}

\usepackage[accsupp]{axessibility}  

\newcommand{\tr}[1]{{#1}^\top}

\newcommand{\mr}[1]{\mathrm{#1}}

\newcommand{\mypar}[1]{\vspace{2pt}\noindent\textbf{#1}~}

\newcommand{\loss}{\mathcal{L}}
\newcommand{\lossSup}{\loss_{\mr{\textsc{ce}}}}
\newcommand{\lossContr}{\loss_{\mr{contr}}}

\newcommand{\lossCam}{\loss_{\mr{cam}}}

\newcommand{\data}{\mathcal{D}}
\newcommand{\trk}{\tau}
\newcommand{\normg}{\widetilde{g}}

\def\wacvPaperID{0775} 

\wacvalgorithmstrack   

\wacvfinalcopy 

\ifwacvfinal
\usepackage[breaklinks=true,bookmarks=false]{hyperref}
\else
\usepackage[pagebackref=true,breaklinks=true,colorlinks,bookmarks=false]{hyperref}
\fi

\begin{document}

\title{Camera Alignment and Weighted Contrastive Learning \\for Domain Adaptation in Video Person ReID}


\author{Djebril Mekhazni$^*$,
  ~Maximilien Dufau$^*$,
  ~Christian Desrosiers,
  ~Marco Pedersoli,
  ~Eric Granger\\
  	LIVIA, Dept. of Systems Engineering, ETS Montreal, Canada\\
{\tt \{djebril.mekhazni, maximilien-francois.dufau.1\}@ens.etsmtl.ca,}
\\
{\tt \{christian.desrosiers, marco.pedersoli, eric.granger\}@etsmtl.ca}
}



\maketitle
\def\thefootnote{*}\footnotetext{These authors contributed equally to this work}\def\thefootnote{\arabic{footnote}}
\begin{abstract}
    Systems for person re-identification (ReID) can achieve a high accuracy when trained on large fully-labeled image datasets. However, the domain shift typically associated with diverse operational capture conditions (e.g., camera viewpoints and lighting) may translate to a significant decline in performance. This paper focuses on unsupervised domain adaptation (UDA) for video-based ReID -- a relevant scenario that is less explored in the literature.  In this scenario, the ReID model must adapt to a complex target domain defined by a network of diverse video cameras based on tracklet information. State-of-art methods cluster unlabeled target data, yet domain shifts across target cameras (sub-domains) can lead to poor initialization of clustering methods that propagates noise across epochs, 
    thus preventing the ReID model to accurately associate samples of same identity.
    In this paper, an UDA method is introduced for video person ReID that leverages knowledge on video tracklets, and on the distribution of frames captured over target cameras to improve the performance of CNN backbones trained using pseudo-labels. 
    Our method relies on an adversarial approach, where a camera-discriminator network is introduced to extract discriminant camera-independent representations, facilitating the subsequent clustering. 
    In addition, a weighted contrastive loss is proposed to leverage the confidence of clusters, and mitigate the risk of incorrect identity associations.
    Experimental results obtained on three challenging video-based person ReID datasets -- PRID2011, iLIDS-VID, and MARS -- indicate that our proposed method can outperform related state-of-the-art methods. 
    Our code is available at: \url{https://github.com/dmekhazni/CAWCL-ReID}
        
\end{abstract}
\section{Introduction}
        Person re-identification (ReID) is an important computer vision function aiming to match individuals captured over distributed network of non-overlapping video cameras. It has a growing interest in real-world applications, such as video surveillance, search and retrieval, and pedestrian tracking for autonomous driving. The non-rigid structure of human bodies, variations in capture conditions in the wild, due to changes in illumination, motion blur, resolution, pose, view point, and occlusion and background clutter leads to high intra-person and low inter-person variability. Despite these considerable challenges, recent progress in deep learning (DL) has allowed to develop state-of-art person ReID models that can achieve an impressive level of performances, especially then they are trained end-to-end in a supervised fashion on large fully-annotated image datasets \cite{TCLNet}.

Given the increasing demand in surveillance and camera networks, video-based ReID has received much attention in recent years.  Video-based ReID systems rely on tracklets -- a consecutive set of bounding box images captured from the same individual -- for person matching, rather than single image. This approach may provide a richer representations by leveraging the temporal coherence of tracklets, allowing for temporal modeling to, e.g., attention-based aggregation of the more important features over time. This reduces the volume of data that must be stored  for person ReID, and the complexity for similarity matching \cite{sekh2020person}. 

The distribution of image data captured with different cameras and operational conditions may differ considerably, a problem known as domain shift~\cite{DMMD}. Given this problem, state-of-the-art DL models that undergo supervised training with a labeled image dataset (from the source domain) often generalize poorly for images captured in a target operational domain, leading to a decline in ReID accuracy.
Moreover, labeled data are costly to obtain especially in ReID due to the nature of the task where unknown and new identities are captured very often. Unsupervised Domain Adaptation (UDA) focuses on transferring the knowledge from a labeled source ($S$) dataset to an unlabeled target ($T$) dataset. It seeks to resolve the domain shift problem by leveraging unlabeled data from the target domain, in conjunction with labeled source domain data, to bridge the gap between the different domains. Most UDA methods for pair-wise similarity matching have been proposed for image-based person ReID, and adopt approaches for domain-invariant feature learning \cite{CORALSun,MMDGANLi,MMDweight,DMMD,NguyenMeidine2021WACV, remigereau2022knowledge},  adversarial training \cite{ECNZhong,zhong2018generalizing,GANWei,ganin2015unsupervised,bousmalis2016domain,tzeng2017adversarial}, and clustering \cite{SPCL}.  
Recently work on adversarial multi-source and multi-target UDA \cite{shen2022benefits,InformationTheoricReg, NguyenMeidine2021WACV} has spurred a growing interest in leveraging the diverse data from individual domains or sub-domains (e.g., video cameras). For instance, the typically apply extended multi-domain adversarial classifiers with information-theoretic weighting strategies allow to regulate training.
 
Given the high cost of annotation, video-based ReID has mostly focused on adaptation with an unsupervised setup, where the source dataset is not available. Clustering and noisy pseudo-labels are used to train the ReID network with no target ground-truth labels \cite{TAUDLchen2018deep,UTALli2019unsupervised,UGAwu2019unsupervised,BUClin2019bottom}. To the best of our knowledge, the UDA problem has not been addressed for video-based person ReID. However, given the complexity of this problem, leverage knowledge provided by video tracklets, or by a small quantity of labeled data may be extremely valuable.  
We observed that the clustering strategy may consequently require tuning of its hyper-parameters \cite{BUClin2019bottom} due to a complex feature space at the initialization. Indeed, a misalignment between the data distribution of target domains cameras at initialization can prevent convergence of the training process. 
Inspired by the clustering and contrastive loss in \cite{SPCL}, we propose to improve the training process on pseudo-labels by improving the alignment of camera data before initialization of pseudo-labels with an adversarial approach. In addition, to mitigate the impact of noisy pseudo-labels during training, we propose a leveraging information from video tracklets, and novel soft contrastive loss. An aspect with these new techniques, and that is not exploited in the literature, is the use of meta-information, such as the camera label and the atomicity of tracklets, i.e, the fact that all the images in a tracklet have one same identity. That were not used before in the literature and is an observation essential for video person ReID.

Our contributions can be summarized as follows. (1) An adversarial camera-level confusion loss is introduced to address the shift in data distributions observed across individual cameras in the target domain. (2) An approach for association of pseudo-labels with the positive pairs (video clips), build from sub-parts of tracklets. These strong positive pairs provide a meaningful heuristic to the clustering algorithm. (3) A soft contrastive loss with dynamic weighting is proposed that relies on a $k$-Nearest Neighbors weighing strategy and video clip-based augmentation to learn discriminant tracklet representations. (4) An extensive set of experiments conducted on three challenging datasets for video-based ReID datasets, PRID2011 \cite{PRID2011}, iLIDS-VID \cite{iLIDS-vid} and MARS \cite{zheng2016mars}, that show our proposed approach outperforming related state-of-the-art models.

    
     

    \section{Related Work}

\subsection{Video-Based ReID}
    In video based person ReID, input samples are video-sequences of the same identity (tracklets). The main objective is to find a good representation of the whole tracklet in order to use it for ReID. This representation is richer than the one extracted from a single image as the temporal information is exploitable, as well as more features can be extracted than in the single image case. The exploitation of this new information required the development of spatio-temporal aggregator networks.
        
    \noindent \textbf{RNNs:} First attempts of aggregation were mostly via Recurrent Neural Networks (RNNs) \cite{mclaughlin2016recurrentRNN,chung2017twoTWOSTREAM,RNNRevisited,RecFeatAgg,ForestTreesSTRec}. It consists in aggregating frame-level features for each individual. Temporal context is stored from previous frames in the hidden state of the RNN. Mclaughlin et al. \cite{mclaughlin2016recurrentRNN} proposed a RNN architecture optimizing both final recurrent layer and temporal pooling layer. A weighted version of it was proposed by Chung et al. \cite{chung2017twoTWOSTREAM}.
     
    \noindent \textbf{Attention Networks:} Another approach for aggregation is to use Attention networks, using temporal information to select the most discriminating frames. While Liu et al. \cite{QAN} consider only temporal attention,  Li et al. \cite{DRSA} and Fu et al. \cite{STA} combine spatial and temporal attention. Hou et al. \cite{TCLNet} differ from previous approaches avoiding redundancy on features of different frames by designing a temporal complementary learning mechanism. They proposes an integral characteristic of each identity by extracting complementary features for consecutive frames of a video. These works leverage the temporal information brought by the video to build richer representations. However these techniques do not leverage video-specific meta-information such as the fact that all the persons appearing in a single tracklet belong to the same identity. Moreover, those methods do not address the problem of domain shift.
 
\subsection{Unsupervised Domain Adaptation for ReID}
    Unsupervised Domain Adaptation rely on transferring knowledge from a labeled source domain to an unlabeled target domain. The goal is to learn discriminant and domain-invariant representations from both domains. 
    Despite the great results achieved by DL methods on supervised person ReID, a domain-shift between source and target data is often present due to the difference across domains (pose, viewpoint, illumination, ...). This leads models trained on source to a drastic performance drop when deployed on target. 
    Different approaches are proposed to address this problem, separated into: domain-invariant feature learning and clustering based.
    \newline
    \textbf{Domain-invariant feature learning:}
        Discrepancy-based approaches measure the shift between source and target distribution. Their objective is to minimize the loss characterising the domain-shift gap, trying to align the confused target distribution on the well separated source distribution \cite{CORALSun,MMDGANLi,MMDweight,DMMD,NguyenMeidine2021WACV}.
        Methods that generate synthetic labeled data aim to produce similar images on source and target data \cite{ECNZhong,zhong2018generalizing,GANWei}.
        However results can vary depending on the quality of the image generation.
        Adversarial-based methods are designed to train a domain discriminator whose goal is to fool the feature extractor using a gradient reversal layer \cite{ganin2015unsupervised,bousmalis2016domain,tzeng2017adversarial}. This allows the feature extractor to make abstraction of the origin of the domain. A supervised loss is present during training not to forgot the initial task and avoid collapse of the representation. 
    \newline
    \textbf{Clustering:}
        Clustering-based approaches work by clustering unlabeled target data and assign them pseudo-labels. Then those labels are used to train the network with a self-supervised method such as the minimization of a contrastive loss defined with pseudo-positives and negatives pairs \cite{BUClin2019bottom,TAUDLchen2018deep,UTALli2019unsupervised,SPCL}. The main issue with this family of approaches is about the quality of the pseudo-labels. Indeed, noise from false positive or false negative pairs could lead to a diverging training. In this context, some criteria are designed to evaluate the coherence of clusters \cite{SPCL}. 
        Contrastive Learning typically requires large batch size to reach satisfying performances. However, the batch size is limited by the size of the GPU memory available. This is particularly important in video-based ReID where each representation is created from multiples images, thus requiring a great amount of memory during training.
        \cite{MemoryBankWu,MoCoHe} allow to contrast samples with the whole dataset by storing in memory all the representation calculated at the beginning of each epoch and then update the network batch-wise. He et al. \cite{MoCoHe} extend this idea with a momentum update after each batch to follow the update by the back-propagation of selected samples.
     \newline
    \textbf {Adversarial Approaches for Multi-Source UDA:}
    Shen et al. \cite{shen2022benefits} proposed a source-free multi-source free DA approach. They estimate the importance of each pretrained model on source to conduct the adaptation onto the target domain incrementally. Park and Lee \cite{InformationTheoricReg} proposed another adversarial approach in the context of multi-source domain adaptation based on classifiers. They dealt with the computational scalability, and the very sensitive gradient variations presents in this kind of approach through a information-theoretic method.
        
\subsection{Self-paced learning}
    Self-paced learning \cite{SPL2010} refers to add progressively more difficult samples across the epochs of the training. It requires to be able to estimate a confidence score and it can help the convergence of the training. In case of semi-supervised or unsupervised learning in a closed-set environment \cite{wang2021self,shen2022benefits} it is possible to base the evaluation on the output of a classifier to work with its probabilities. Also we need to differentiate the weighting of domains in case of multiple source or target \cite{shen2022benefits} which is different and does not apply in our case.
    However, it is not possible to obtain this value in the context of the unsupervised open-set problem. In this case, identities between source and target are totally different and we do not know the number of classes also. As mentioned above, clustering is very noise-sensitive.  In person ReID, Ge et al. \cite{SPCL} proposed two criteria to manage the clustering based on tightness and compactness.  In contrast, we propose to assimilate a weight to pairs in the contrastive loss directly, allowing less negative impact in case of bad associations.
        
\subsection{Critical Analysis}

    Although unsupervised ReID has been widely studied recently, most of the works were designed in an image-based scenario, not taking advantage of temporal features \cite{SPCL,ECNZhong,GANWei} yet full of information inside a tracklet specific to its identity.
    Moreover, in this work we are the first at our best knowledge who consider the camera alignment issue for a given domain, which can lead to a poor representation of the tracklets. As clustering-approaches are very noise-sensitive, the camera alignment is really important for good results and we propose an adversarial loss to fulfill it.
    Finally, few existing works in unsupervised video-based ReID leveraged the information of a source dataset in a video context. 
    In a more general context, video-based approaches are missing a more comprehensive study regarding the importance of the tracklet \cite{BUClin2019bottom,TAUDLchen2018deep,UTALli2019unsupervised}.
    Although the development of pertinent features is an important point to be approached, managing in the first place the raw data across tracklets requires still some efforts to fully benefit the hidden information available from the raw data.


    \section{Proposed Approach}
         \label{sec:methodology}
\begin{figure*}[!ht]
    \centering
    \includegraphics[width=0.8\textwidth]{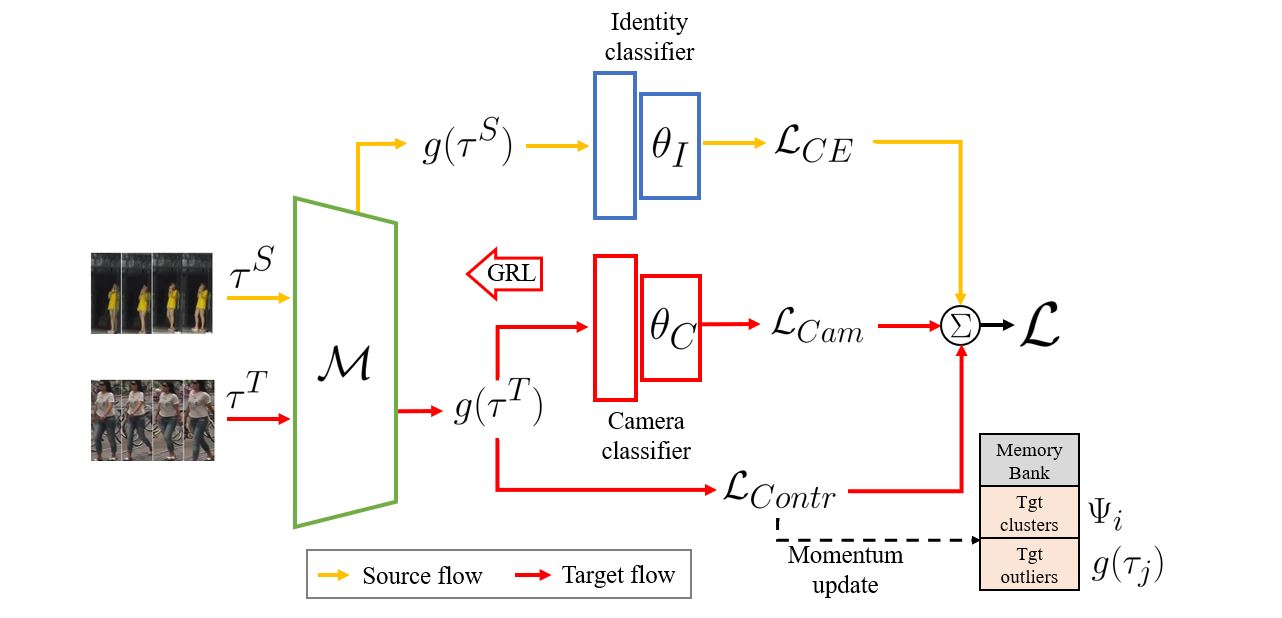}
    \caption{Training architecture of the proposed model for video-based UDA in ReID. Source and target tracklets are processed by the backbone $\mathcal{M}$ and a representation $g(Tk)$ is computed. The training loss $\mathcal{L}$ is composed of three part: the supervised cross-entropy loss on the source tracklets ($\mathcal{L}_{CE}$), camera confusion loss on source and target tracklets ($\mathcal{L}_{Cam}$) and cotrastive loss ($\mathcal{L}_{Contr}$), which uses a memory bank of clustered and unclustered tracklets. }
    \label{fig:overall_framewotk}
\end{figure*}

 In this section, the overall approach shown in Figure \ref{fig:overall_framewotk} is presented module by module.
Most methods for video person ReID learn to associate tracklets to the correct identity through a clustering approach based on pseudo-labels.  
However, those method strongly rely on the quality of the pseudo-labels in the early training iterations. Poor initialization will prevent iterative convergence and the whole training process will fail. We propose a novel training process based on domain adaptation that starts by consolidating the performances of the encoder on the Target dataset with an Adversarial Camera confusion technique. After improving the cluster quality thanks to previous step, the network can start the self-paced pseudo-label based training. Self-paced is essential to estimate bad clusters and avoid encouraging wrong directions during the training. From our best knowledge, we are the first using the camera ID information to decrease the encoder discrimination on the Target set.
    
\subsection{Tracklet Representation}

    Given a set $\data$ of $N$ tracklets, each tracklet is composed of a sequence of $N_\trk$ images $\{x_1, \ldots, x_{|N_\trk|}\}$. We consider that the first $N_s < N$ tracklets are from a source domain $S$ and come with a personID label $y$, while the remaining $N_t = N-N_s$ ones are from a target domain for which there is no person ID. The goal is to learn a function $f$ mapping images $x$ to a representation vector $f(x)$ which is able to discriminate identities, i.e. find separable distances to consider two pairs as same individual or no, while being robust to differences between the source and target domains. 
    
    Since input data are tracklets, a features aggregator is employed so each video-sequence has its own unique representation.  The aggregation method TCLNet proposed by Hou \cite{hou2020temporal} is adopted. By exploiting spatio-temporal information, TCLNet is the association of Temporal Saliency Erasing (TSE) and Temporal Saliency Boosting (TSB) modules. 
    We denote the backbone $\mathcal{M}$ as the layer just before the classification layer. It produces the representation of a tracklet $g(\trk)$.
    \newline
    \textbf{Clips:} In the video-based context, it happens that tracklets contain different mode. Between the starting frame and the final frame, events could have arrived (occlusion, change in trajectory, ...). From this observation, we proposed to split each tracklet $\trk$ into $N_\Omega$ equal-sized clips $\Omega$. The tracklet becomes then $\trk = \{\Omega_0, ..., \Omega_{Nc} \}$


\subsection{Training Strategy}
    An association of a supervised loss, a camera confusion loss and a self-paced contrastive learning loss for domain adaptation is proposed as follow:    
        \begin{equation}
        \loss \, = \, \delta_1\lossSup \, \, + \, \delta_2\lossCam \, + \delta_3\lossContr.
        \end{equation}
    \noindent where $\lossSup$ refers to the supervised training on source = S domain, $\lossCam$ refers to the camera classifier to avoid learning camera-identity related, and $\lossContr$ refers to the contrastive loss on clustered samples. $\delta_i$ are weighting factors used to balance the magnitude of the losses.

\subsection{Supervised training}

    The first loss, which uses only source data, ensures that the tracklet representation $g(\trk)$ can be used for identity classification through an identity classifier $\theta_I$. It converts $g(\trk)$ to a class probability vector with a linear layer followed by a softmax, and uses cross-entropy as classification loss. It can be expressed as:
    \begin{equation}
        \mathcal{L}^{\mr{\textsc{ce}}} =
        \frac{1}{K}\sum_i^K
        -\log \left( \frac{ \exp ( {W^T_{y_i} g(\trk_i) + b_{y_i}} ) } {\sum_{j=1}^{N_I} \exp{ ( W_{j}^T g(\trk_i) + b_j} ) }   \right)
    \end{equation}
    \noindent where $K$ is the batch size.  Class label $y_{i} \in \{1, 2, ..., N_I\}$ is associated with training tracklet g($\trk_i$), the $i^{th}$ training tracklet. Weight vectors $W_{yi}$ and bias $b_{yi}$ of last fully connected (FC) layer corresponds to class $y$ of $i^{th}$ tracklet. $W_{j}$ and $b_{j}$  are weights and bias of last FC corresponding of the $j^{th}$ class ($j \in [1, N_I]$). $W_j$ and $W_{y_i}$ are respectively the $j^{th}$ and $y_i^{th}$ column of $W = \{ w_{ij} : i = 1, 2, ..., F; j = 1, 2, ..., N_I\}$, where $F$ is the size of the last FC layer of the backbone.

\subsection{Camera-level confusion loss}

    \label{clc}  
    We empirically observed a mis-alignment of target = T inter camera distributions after the pre-training on the Source dataset. This lack of alignment yield poor initial pseudo-labels.  Inspired by \cite{ganin2015unsupervised} we use an adversarial distribution alignment approach. We propose to align the per-camera distribution rather than the domains. A camera classifier $\theta_c$ is trained to discriminate the target = T cameras with a cross entropy loss, the gradient of this loss is inverted with a Gradient Reversal Layer (GRL) and back-propagated through the encoder. This reversed objective aims to remove the information related to the camera scene from the representation and helps to make the source and target domains indistinguishable for the classifier. Using the cross-entropy loss to train the camera classifier incentives the encoder to minimize the classifier accuracy, which can be unstable.
    \begin{equation}
        \loss^{\mr{\textsc{ce}}}_{\mr{cam}}     
        \frac{1}{K}\sum_i^K
        -\log \left( \frac{ \exp ( {W^T_{c_i} \theta_c(g(\trk_i)) } ) } {\sum_{j=1}^{N_c} \exp{ ( W_{j}^T \theta_c(g(\trk_i)) } ) }   \right)
        \label{eq:CE_camcla}
    \end{equation}
    \noindent where $N_c$ is the number of cameras $T$ in the target domain.  We therefore propose a novel Camera Level Confusion loss $\loss_{\mr{cam}}^{\mr{\textsc{clc}}}$ to train the camera classifier as:
    \begin{equation}
        \begin{aligned}
         \loss^{\mr{\textsc{clc}}}_{\mr{cam}} =  & \quad
        \frac{1}{K^2} \left[\sum^{K}_{i=1}\log \frac{\exp c_{y_i,i}}{\sum_{j=1}^{N_c}\exp(c_{j,i})}\right].
        \\
        & \quad
        \left[\sum^{K}_{i=1}\log \frac{\exp (    
        1 -
        c_{y_i,i})}{\sum_{j=1}^{N_C}\exp( 1- c_{j,i})}\right]
         \end{aligned}
        \end{equation}
    \noindent where $c_{j,i}$ is the output of the classifier for camera $j$ and sample $i$.        
    This loss, encourages the classifier to be sure about his predictions, may they be right or wrong. The encoder is then encouraged to maximise the classifier confusion, i.e maximise the conditional entropy $H(C|g(\trk))$ where $C$ is the distribution of cameras and the distribution of tracklets. A similar confusion based loss for adversarial approaches was proposed in \cite{InformationTheoricReg}.

\subsection{Contrastive loss}

    The contrastive loss is based on pseudo-positive/negatives pairs of samples assimilated to the same or different clusters from $T$ data. It is defined as:
            \begin{equation}\label{eq:lossContr}
                \lossContr \ = \ \frac{1}{N_t} \sum_{i=1}^{N_t} \sum_{j\neq i} \log \frac{\exp\left(\tr{\normg(\trk_i)} \normg(\trk_j)\right)}
                {\sum\limits_{j' \neq i} \exp\left(\tr{\normg(\trk_i)} \normg(\trk_{j
                '})\right)} 
            \end{equation}
    \noindent where $\normg(\trk) = g(\trk) \mathbin{/} \|g(\trk)\|$. This loss helps to adapt the tracklets representations such that different identities are easily distinguished on both source and target data.
    
    \mypar{1) Generation of Pseudo-Labels:} To generate pseudo-labels, DBSCAN clustering algorithm is used. A Jaccard distance matrix, calculated based on the $k$-reciprocal nearest neighbours sets, is used for this clustering phase. The choice of such a metric enforce more robust clusters. Indeed, samples that share reciprocal neighbors tend to form positive pairs \cite{SPCL, RerankingZhong}.
    
    \mypar{2) Self-Paced Strategy:} Let $w_{ij} \in [0,1]$ be a self-paced weight representing the likeliness that tracklets $\trk_i$ and $\trk_j$ correspond to the same person \cite{khosla2020supervised}. 

    For target examples, we need to optimize both embedding function $f$ and self-paced weights $w_{ji}$. The update of self-paced weights is controlled by a self-paced regularization function $R_\gamma$, parameterized by the learning pace $\gamma > 0$. This function has the property of being monotone decreasing with respect to the loss (i.e., harder examples are given less importance) and monotone increasing with respect to the learning (i.e., a larger $\gamma$ increases the weights). 
    Here, we consider for this function a linear self-paced learning:    
        \begin{equation}
            R_\gamma(w_{ij}) = -\gamma w_{ij}.
        \end{equation}       

    Given the self-paced contrastive loss (\ref{eq:lossContr}):
        \begin{equation}
            \lossContr \ = \ \frac{1}{N_t} \sum_{i=1}^{N_t} \sum_{j\neq i} w_{ij}  l_{ij} \, + \, R_\gamma\big(w_{ij}\big) \, \mbox{, where}
        \end{equation}
        \begin{align}\label{eq:lij}
            l_{ij} 
            & =  - \log \frac{\exp\left(\tr{\normg(\trk_i)} \normg(\trk_j)\right)}
            {\sum\limits_{j' \neq i} \exp\left(\tr{\normg(\trk_i)} \normg(\trk_{j
            '})\right)}
        \end{align}
    Since $l_{ij} \geq 0$ and $w_{ij} \in [0,1]$, $w$ can be optimized in closed-form for a fixed embedding function $g$ as:
        \begin{align}
            w_{ij} =  \max\Big(1-\frac{1}{\gamma}l_{ij}, \, 0\Big).
        \label{eq:sp-weights}
        \end{align}        
         
    
    The overall optimization process starts with a small learning rate $\gamma_0 \approx 0$, at each iteration, we update the embedding function $g$ gradient descent and updating the weights $w_{ij}$ as in (\ref{eq:sp-weights}). At the end of each iteration, the learning pace can be increased according to a fixed policy, for instance $\gamma := (1+\alpha)\gamma$, with $\alpha > 0$. In our method, we set $\gamma$ based on the maximum distance between an example and its $k$-nearest neighbor. This policy treats $k$-nearest neighbors as partially positive pairs and thus mitigates the harm of false-negative close pairs. Indeed, close pseudo-negative pairs are more likely to actually be positive pairs.
        
    \mypar{3) Memory bank:} To alleviate memory issues and allow the full use of all data for the contrastive loss, allowing a more reliable comparison pairs to pairs at each batch step, a memory bank is employed to store the features of all samples at the beginning of each epoch.
    Clustered samples have the mean of the features belonging to the same cluster $\Psi_i$, with i the i-th cluster, saved in the memory. For unclustered samples we save their features $g(\trk_j)$ directly, with j the j-th sample unclustered.
    It consist into considering a query q and a set of encoded samples \{$k_0, k_1, ...$ \} that are keys of a dictionary. The contrastive loss \ref{eq:lossContr} is then calculated considering samples in the batch as q and all other samples store in the dict as keys. The backpropagation and weight update for the encoder are made only with samples in the batch, not including the keys being detached from the graph. To avoid memory issues in storing all the graph for them, we update the keys differently, according to the following rule \cite{MoCoHe}:
            \begin{equation}
                \theta_k \leftarrow m\theta_k + (1 - m)\theta_q.
            \end{equation}
    In this context, $m \in [0, 1]$ and $\theta_k$ (resp. $\theta_q)$ refers to the representation of the keys (resp. query).
    
    \section{Results and Discussion}
        \subsection{Experimental Methodology}

    The proposed method is evaluated on the three most challenging public video re-identification benchmarks.
    
    PRID2011 \cite{PRID2011} is the least challenging of the three benchmarks, it features 356 videos from 178 identities captured by two cameras monitoring crosswalks and sidewalks. iLIDS-VID \cite{iLIDS-vid} is captured in an airport hall. It is a two camera benchmark with 600 videos from 300 identities. This benchmark is challenging due to pervasive background clutter, mutual occlusions, and lighting variations. MARS \cite{zheng2016mars} is the current largest open video-based pedestrian retrieval benchmark. It features six cameras on a university campus, for a total of 20,478 videos from 1,261 identities.
        
    \begin{table}[ht]
    \centering
    \resizebox{\columnwidth}{!}{
        \begin{tabular}{l|ccc}
            \toprule
            \bf Property
            & \textbf{MARS}              & \textbf{PRID2011}           & \textbf{iLIDS}          \\  \midrule
            \# Identities                                               & 1251              & 178                 & 300                \\ 
            ID Train/Query/Gallery                                      & 625/626/622       & 89/89/89            & 150/150/150        \\ 
            Train/Query/Gallery                                         & 8298/1980/9330    & 178/89/89           & 300/150/150        \\ 
            \# Tracklets                                                & 19608             & 356                 & 600                \\ 
            \# Image per Tracklet                                       & 2$\sim$920 (59.5) & 28$\sim$675 (108.1) & 22$\sim$192 (70.8) \\ 
            \# Cameras                                                  & 6                 & 2                   & 2                  \\ 
            \# Images                                                   & 1,191,003         & 24,541               & 42,495              \\ 
            Annotation Method                                                & DPM+GMMCP         & Hand                & Hand               \\ 
            Crop Size                                                   & 256x128           & 128x64              & Vary               \\ 
            \bottomrule
        \end{tabular}
    }
\caption{Description of the different datasets used in experiments -- MARS, PRID2011 and iLIDS-VID.}
\label{table:database-video}
\end{table}
    
    The proposed method is implemented using PyTorch. We use ResNet50 \cite{ResNet50} pretrained on ImageNet as the backbone of the spatio-temporal aggregator TCLNet \cite{hou2020temporal}. We separate each video in $K=2$ clips, from which we randomly select four images to form our input tracklet.  We train the model for 150 epochs with batch size 24. Adam optimizer is used to perform gradient descent with an initial learning rate of 0.0003, a scheduled decay at epochs 40, 80 and 120 and weight decay parameter $\gamma$ equal 0.1. $\delta_1$ is set at 1, $\delta_2$ is set at 0.2, and $\delta_3$ at 1.
    At evaluation time the features are $\ell_2$ normalized. We use the $\ell_2$ distance between the gallery and query set for the final matching.
    For the clips approach, we proposed to split the tracklet into N parts by taking N times successive frames from the beginning to the end. In our case N is set at 2 for best results as stated Table \ref{tab:number_clips}, meaning the first 50\% frame of each tracklet are the first clip and the rest the second clip. Then each clips is separated by a fixed number of equal chunks (4 in our case) and randomly take a frame from each chunks. We believe that it can be used for video-based person ReID and even be extended for Video-based tasks such as action recognition as an example.

\subsection{Ablation Study}

    \small{\begin{table*}[!ht]
    \centering
    \begin{small}
    \begin{tabular}{l|ccc}    
        \toprule
        \multirow[b]{1}{*}{ \textbf{Loss Functions} } & 
        \multicolumn{1}{c}{\begin{tabular}[c]{@{}c@{}} iLIDS $\rightarrow$ PRID \\ (2 cameras)
        \end{tabular}} & 
        \multicolumn{1}{c}{\begin{tabular}[c]{@{}c@{}} PRID $\rightarrow$ iLIDS \\ (2 cameras)\end{tabular}} & 
        \multicolumn{1}{c}{\begin{tabular}[c]{@{}c@{}} iLIDS $\rightarrow$ MARS \\ (6 cameras)\end{tabular}} \\  \midrule
        Lower Bound (sup. $S$ only) & 49.0 & 12.7 & 19.8 \\ 
        \midrule        
        \multicolumn{1}{l|}{
        \begin{tabular}[l]{@{}l@{}} $\mathcal{L}_{\mr{cam}}^{\mr{\textsc{ce}}}$ \\ $\mathcal{L}_{\mr{cam}}^{\mr{\textsc{clc}}}$ \\ $\mathcal{L}_{\mr{cam+clips}}^{\mr{\textsc{clc}}}$\end{tabular}}              & \multicolumn{1}{c}{\begin{tabular}[c]{@{}c@{}}61.8\\ 66.3\\ 75.3 \end{tabular}}          & \multicolumn{1}{c}{\begin{tabular}[c]{@{}c@{}}32.0\\ 35.3\\ 36.0 \end{tabular}} & \multicolumn{1}{c}{\begin{tabular}[c]{@{}c@{}}30.8\\ 31.5\\ 32.3 \end{tabular}}          \\ \midrule        
        \multicolumn{1}{l|}{\begin{tabular}[c]{@{}l@{}}$\mathcal{L}_{\mr{contr}}$\\ $\mathcal{L}_{\mr{contr}}^{\mr{kNN}}$\\ $\mathcal{L}_{\mr{contr+clips}}^{\mr{kNN}}$\end{tabular}}                   & \multicolumn{1}{c}{\begin{tabular}[c]{@{}c@{}}77.6\\ 78.2\\ 79.8\end{tabular}}          & \multicolumn{1}{c}{\begin{tabular}[c]{@{}c@{}}41.9\\ 44.0\\ 44.7\end{tabular}} & \multicolumn{1}{c}{\begin{tabular}[c]{@{}c@{}}37.6\\ 38.5\\ 38.9\end{tabular}}          \\ \midrule        
        \multicolumn{1}{l|}{\begin{tabular}[l]{@{}l@{}}
        $\loss_{\mr{contr}}$ + $\loss_{\mr{cam}}^{\mr{\textsc{clc}}}$\\ $\loss_{\mr{contr}}^{\mr{kNN}}$ + $\loss_{\mr{cam}}^{\mr{\textsc{clc}}}$\\ $\loss_{\mr{contr+clips}}^{\mr{kNN}} + \loss_{\mr{cam+clips}}^{\mr{\textsc{clc}}}$\end{tabular}} & \multicolumn{1}{c}{\begin{tabular}[c]{@{}c@{}}83.9\\ 84.2\\ \textbf{86.5} \end{tabular}} & \multicolumn{1}{c}{\begin{tabular}[c]{@{}c@{}}54.0\\ 58.0\\ \textbf{58.3} \end{tabular}} & \multicolumn{1}{c}{\begin{tabular}[c]{@{}c@{}}59.6\\ 60.2\\ \textbf{62.2} \end{tabular}} \\ \bottomrule        
    \end{tabular}
    \end{small}
    \caption{Rank-1 accuracy with different losses functions for the camera classifier and contrastive learning, and when subdividing tracklets into 2 clips (best result obtained in Table \ref{tab:number_clips}).}
    \label{tab:ablation}
\end{table*}
}

    \small{\begin{table*}[!ht]
    \centering
    \begin{small}
    \begin{tabular}{lc|cccccc}
        \toprule
        \multirow[b]{2}{*}{ \textbf{Method} } & 
        \multirow[b]{2}{*}{ \textbf{Setting} } & 
        \multicolumn{2}{c}{\begin{tabular}[c]{@{}c@{}} iLIDS $\rightarrow$ PRID \\ (2 cameras)\end{tabular}} & 
        \multicolumn{2}{c}{\begin{tabular}[c]{@{}c@{}} PRID $\rightarrow$ iLIDS \\ (2 cameras)\end{tabular}} & 
        \multicolumn{2}{c}{\begin{tabular}[c]{@{}c@{}} iLIDS $\rightarrow$ MARS \\ (6 cameras)\end{tabular}} \\ 
        \cmidrule(l{5pt}r{5pt}){3-4}\cmidrule(l{5pt}r{5pt}){5-6}\cmidrule(l{5pt}r{5pt}){7-8}
        &  & \multicolumn{1}{c}{Rank-1}  & mAP & \multicolumn{1}{c}{Rank-1}   & mAP    & \multicolumn{1}{c}{Rank-1} & mAP  \\ \midrule
        Lower Bound (sup. $S$ only) & --   & \multicolumn{1}{c}{49.0} & 60.0  & \multicolumn{1}{c}{12.7}                     & 20.9                                   & \multicolumn{1}{c}{19.8}                           & 10.1      \\ \midrule
        DGM+IDE \cite{DGMye2017}, ICCV'19   & OneEx      & \multicolumn{1}{c}{--}                        & \multicolumn{1}{c}{--}          & \multicolumn{1}{c}{--}                         & \multicolumn{1}{c}{--}                  & \multicolumn{1}{c}{36.8}                           & 16.8     \\
        Stepwise \cite{SMP}, ICCV'17       & OneEx      & \multicolumn{1}{c}{--}                        & \multicolumn{1}{c}{--}             & \multicolumn{1}{c}{--}                        & \multicolumn{1}{c}{--}                & \multicolumn{1}{c}{41.2}                           & 19.6   \\
        EUG \cite{EUG}, CVPR'18            & OneEx   & \multicolumn{1}{c}{--}                         & \multicolumn{1}{c}{--}             & \multicolumn{1}{c}{--}                         & \multicolumn{1}{c}{--}                 & \multicolumn{1}{c}{\textbf{62.2}}                           & 42.5   \\ \midrule
        TAUDL \cite{TAUDLchen2018deep}, BMVC'18    & Unsup                    & \multicolumn{1}{c}{85.3}                      & --                                 & \multicolumn{1}{c}{56.9}                     & --                                      & \multicolumn{1}{c}{46.8}                           & 21.4       \\ 
        UTAL \cite{UTALli2019unsupervised}, TPAMI'19  & Unsup                    & \multicolumn{1}{c}{54.7}                      & --                                 & \multicolumn{1}{c}{35.1}                     & --                                      & \multicolumn{1}{c}{49.9}                           & 35.2    \\ 
        UGA \cite{UGAwu2019unsupervised}, ICCV'19  & Unsup                    & \multicolumn{1}{c}{80.9}                      & --                                 & \multicolumn{1}{c}{57.3}                     & --                                      & \multicolumn{1}{c}{58.1}                           & 39.3    \\ 
        BUC \cite{BUClin2019bottom}, AAAI'19  & Unsup                    & \multicolumn{1}{c}{--}                        & --                                 & \multicolumn{1}{c}{--}                       & --                                     & \multicolumn{1}{c}{61.1}                           & 38.0       \\ 
        Soft Sim. \cite{SoftSimlin2020unsupervised}, CVPR'20  & Unsup                    & \multicolumn{1}{c}{--}                        & --                                 & \multicolumn{1}{c}{--}                       & --                                      & \multicolumn{1}{c}{61.9}                           & 43.6    \\ \midrule
        SPCL* \cite{SPCL}, NeurIPS'20   & UDA                     & \multicolumn{1}{c}{77.6}                      & 82.1                              & \multicolumn{1}{c}{41.9}                     & 47.6                                   & \multicolumn{1}{c}{37.6}                           & 20.4    \\ 
        Ours ($\loss_{\mr{cam}}^{\mr{\textsc{clc}}}$)           & UDA                     & \multicolumn{1}{c}{70.8}                      & 77.3                              & \multicolumn{1}{c}{32.0}                     & 42.6                                   & \multicolumn{1}{c}{31.5}                           & 16.3    \\ 
        Ours ($\loss_{\mr{cam}}^{\mr{\textsc{clc}}}$ + $\loss_{\mr{contr}}^{\mr{kNN}}$) & UDA      & \multicolumn{1}{c}{\textbf{86.5}}                      & \textbf{89.9}                              & \multicolumn{1}{c}{\textbf{58.3}}                     & \textbf{66.7}                                   & \multicolumn{1}{c}{\textbf{62.2}}                           & \textbf{44.8}     \\ \midrule
        
        Upper Bound (sup. $S \cup T$)  & Tuning  & \multicolumn{1}{c}{92.1}  & 94.5                              & \multicolumn{1}{c}{76.0}                     & 84.0                                   & \multicolumn{1}{c}{86.9}                           & 81.8        \\ 
        \bottomrule
    \end{tabular}
   \end{small}
    \caption{Accuracy of the proposed and state-of-art methods when alternating iLIDS, PRID and MARS datasets as sources target domains ($S \rightarrow T$). Lower bound performance is obtained through supervised training on the labeled source dataset only, while the upper bound after supervised fine-tuning on both source and target data. ``OneEx'' (One Example) settings indicates approaches that labeled one tracklet per identity. ``Unsup'' (Unsupervised) is for end-to-end setting without using source data, while UDA involves using source data. *SPCL was designed for image-based person ReID, while unsupervised methods were designed considering temporal information in tracklets.}
    \label{tab:SOTA}
\end{table*}
}

    In this section, we assess the impact on performance of different losses for the camera classifier and contrastive learning, and the use of clips for the video tracklets.
    
    \mypar{1) Adversarial Adaptation for Camera-Wise Domain Shift:} Table \ref{tab:ablation} compare the two proposed losses for the camera classifier $\loss_{\mr{cam}}^{\mr{\textsc{ce}}}$ and $\loss_{\mr{cam}}^{\mr{\textsc{clc}}}$ in the first section. 
    We observe that the impact on results is significant for all scenario compared to the baseline (deployment of the source trained model on target directly). We achieve 61.8\% (PRID2011), 35.7\% (iLIDS-VID) and 30.8\% (MARS) Rank-1 for $\loss_{\mr{cam}}^{\mr{\textsc{ce}}}$. Even better is our proposed loss $\loss_{\mr{cam}}^{\mr{\textsc{clc}}}$ that improve results by 4.5\% (PRID2011), 3.3\% (iLIDS-VID) and 0.7\% (MARS).
        
\begin{figure}[!t]
        \centering
        \includegraphics[width=0.48\textwidth]{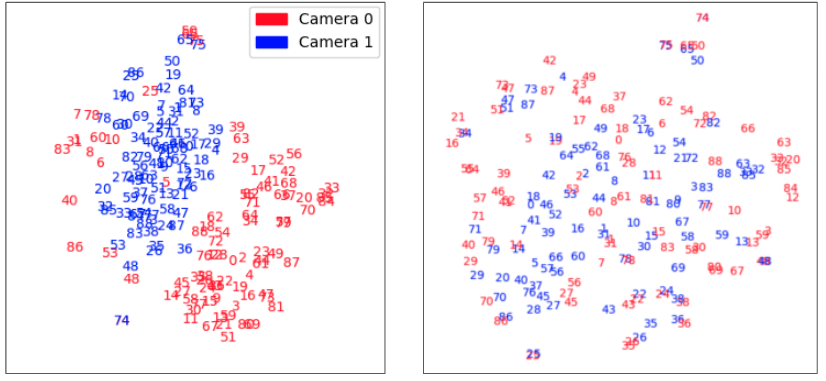}
        \caption{T-SNE plot showing the shift in data distribution among PRID2011 cameras (T) before (left) and after (right) adversarial adaptation. Each number represents an identity.}
        \label{fig:cameraGap}
    \end{figure}

    Figure \ref{fig:cameraGap} illustrates a before and after the distribution shift across cameras of the target domain. We observe that our method successfully shorten the distribution shift between the different cameras on PRID2011. 
    We conducted  experiments using source cameras also, but since the source already associated pairs from different cameras close to perfectly, results were similar or a little bit under because of potential degradation from the camera classifier. Moreover, the introduction of the confusion loss yields a significant improvement in performances. This is a consequence of the more consistent target for the encoder network that aims at maximising the conditional entropy introduced in Section \ref{clc} rather than minimizing the accuracy of the classifier, which can lead to unstable training performances.
    
    \mypar{2) Soft Contrastive Loss:} In the second part of Table~\ref{tab:ablation}, we compare the performance of the initial technique from SPCL \cite{SPCL} applied for Video-based ReID scenario, that we named $\loss_{\mr{contr}}$, and the proposed amelioration $\loss_{\mr{contr}}^{\mr{kNN}}$ where we weight the contrastive loss based on the criterion using k- nearest neighbors.
    We observe that the kNN based weighting improve performances by 0.6\% (PRID2011), 2.1\% (Ilids-Vid) and 0.9\% (MARS) for Rank-1.
     \newline
    \mypar{3) Combination of Losses:} On one side, the contrastive loss \cite{SPCL} train the network based on a self-paced clustering. On the other side, the camera classifier eliminates camera shift observed in Video-based and allow an improvement on results.
    These two components both aim to improve performance by mitigating the harm caused by incorrect pseudo labels in different manners and can be used in combination. The combination of the losses allows the method to improve results by a significant margin as presented in the last section of \ref{tab:ablation}. Especially, the harder the dataset is and the larger the impact of the combination. Indeed, for PRID we only have 6.0\% of improvement compared to Section 2 while we not that for iLIDS we have 14.0\% improvement and 21.7\% for MARS. This indicates the need for the camera classifier, especially for complex domains where the generation of pairs is hard due to domain shift across cameras.
    \newline
    \mypar{4)Video ReID Using Clips:}  Since Video-based person ReID has access to each frame of the whole tracklet, it is possible to split it in clips. We observe in Table \ref{tab:ablation} that dividing each tracklet in clips, as explained in Section \ref{sec:methodology} improves the performance of our method for all loss combinations. It allows the model in its best configuration to reach 86.5\% Rank-1 accuracy for PRID (+ 2.3\%), 58.3\% for iLIDS-VID (+ 0.3\%) and 62.2\% for MARS (+ 2.0\%).
    This can be explained by multiple facts. First, the number of samples in the dataset increases, so it can be seen as data augmentation but without transformation. For smallest dataset it is pertinent because it allow to avoid overfitting. Second, it generates strong reliable positives pairs, which provides a supervised signal even with unlabeled data. From our experiments we observed that splitting a tracklet into two clips is the most promising as stated in Table \ref{tab:number_clips}.
        
        \small{\begin{table}[!ht]
    \centering
    \begin{small}
    \begin{tabular}{c|ccc}
        \toprule
        \bf\#clips & \bf PRID & \bf iLIDS & \bf MARS \\ 
        \midrule
        1        & 84.2 & 58.0  & 60.2 \\
        2        & 86.5 & 58.3  & 62.2 \\
        3        & 86.1 & 58.1  & 61.9 \\
        4        & 85.4 & 58.0  & 60.9 \\
        8        & 83.1 & 56.6  & 59.5 \\
        \bottomrule
    \end{tabular}
    \end{small}
    \caption{Rank-1 accuracy of the proposed method as a function of the number of clips per tracklet.}
    \label{tab:number_clips}
\end{table}
}
        
    \subsection{Comparison with the State-of-Art}
    
        Table \ref{tab:SOTA} compare results to the existing literature for Unsupervised Person ReID \cite{TAUDLchen2018deep,UTALli2019unsupervised,UGAwu2019unsupervised,BUClin2019bottom,DBC,SoftSimlin2020unsupervised,SPCL} have been conducted.
        For more reference we included few shot techniques \cite{DGMye2017,SMP,EUG} where they allow one tracklet per identity to be labeled, giving them a considerable advantage, especially because it informs about the total number of classes.
        In our case, the Domain Adaptation part is present only for the camera classifier where we need to maintain a supervised training to adjust properly the confusion on the backbone, so the backbone is not fully updated via the camera classifier.
        The clusters and the contrastive loss are performed without any source or supervised information.
        For end-to-end approaches, TAUDL \cite{TAUDLchen2018deep} and UTAL \cite{UTALli2019unsupervised} have trouble to perform well on the three proposed dataset. Indeed as example TAUDL perform well on PRID (85.3\% Rank-1 compared to 86.5 \% for ours) but fails when dealing with complex dataset (46.8\% Rank-1 compared to 62.2\% for us).
        BUC \cite{BUClin2019bottom} perform well but is clustering based approach and requires high amount of resources in term of hyper-parameters search. We have a margin of 1.1\% Rank-1 and 6.8\% mAP compared to them. In the same idea is DBC \cite{DBC} but we did not include them because it is hard to reproduce their results.
        Compared to methods with one example setup, we outperform DGM+IDE and Stepwise \cite{DGMye2017,SMP} by a significant margin. We obtain similar results compared to EUG \cite{EUG} for Rank-1 but reach 2.3\% more mAP in our case without any labeled tracklets.
    
    \section{Conclusion}
        This work proposes a novel training approach for the domain adaptation video-based person ReID task, that can be applied to most existing encoders architectures. To solve this challenging task we propose to use the meta-information of the problem to reduce our dependence to the noisy pseudo-labels.
Our method achieves competitive results with state-of-the-art. Moreover, apart from the camera classifier, our ideas aren't specific to the video-based person ReID task and could be use in others time-series representation learning tasks. 
Future direction of the project include improving the inter-Clips diversity with ReID specific augmentation and sampling, as well as improving the clustering with the prior obtained with the usage of clips.


    \section*{Acknowledgments:} 
        This research was supported in part by the Ericsson Global AI Accelerator, Compute Canada and MITACS.

    \newpage
    
    {\small
    \bibliographystyle{ieee_fullname}
    \bibliography{egbib}
    }
    \newpage
    \section{Supplementary Material}
    
        \subsection{Ablation: Impact of Frames per Tracklets}

            \begin{table}[!ht]
                \begin{tabular}{|c|cc|cc|cc|}
                \hline
                \multirow{2}{*}{fpt} & \multicolumn{2}{c|}{PRID2011}      & \multicolumn{2}{c|}{iLIDS-VID}     & \multicolumn{2}{c|}{MARS}          \\ \cline{2-7} 
                                     & \multicolumn{1}{c|}{mAP}  & R-1 & \multicolumn{1}{c|}{mAP}  & R-1 & \multicolumn{1}{c|}{mAP}  & R-1 \\ \hline
                2                    & \multicolumn{1}{c|}{87.2} & 89.6   & \multicolumn{1}{c|}{72.5} & 78.4   & \multicolumn{1}{c|}{81.8} & 75.0   \\ \hline
                4                    & \multicolumn{1}{c|}{90.1} & 92.5   & \multicolumn{1}{c|}{74.0} & 82.0   & \multicolumn{1}{c|}{84.9} & 79.8   \\ \hline
                8                    & \multicolumn{1}{c|}{89.9} & 92.2   & \multicolumn{1}{c|}{74.2} & 82.0   & \multicolumn{1}{c|}{84.6} & 78.6   \\ \hline
                16                   & \multicolumn{1}{c|}{90.1} & 92.9   & \multicolumn{1}{c|}{73.7} & 82.2   & \multicolumn{1}{c|}{85.1} & 79.3   \\ \hline
                \end{tabular}
                \caption{Accuracy for a supervised training using a different number of frames per tracklets (fpt).}
                \label{tab:framespertracklets}
            \end{table}
            
            Table \ref{tab:framespertracklets} analyses the impact on performance of growing the number of frames per tracklet (fpt). A CNN backbone has been trained with the supervised contrastive loss function. Each tracklet is cut into equal size to maintain scalability across a dataset. We can observe that the temporal information remains stable from 4 frames per tracklets, and performances does not differ significantly beyond that point. In this context, setting the number of fpt to the minimal value that provides the best accuracy (4 in our case) is suitable parameter setting for real-time video-based ReID applications. It is less costly than longer tracklets in term of memory consumption.
        
        \subsection{Visual Results}
        
\begin{figure*}[btp]
                 \centering
                 \includegraphics[width=\textwidth]{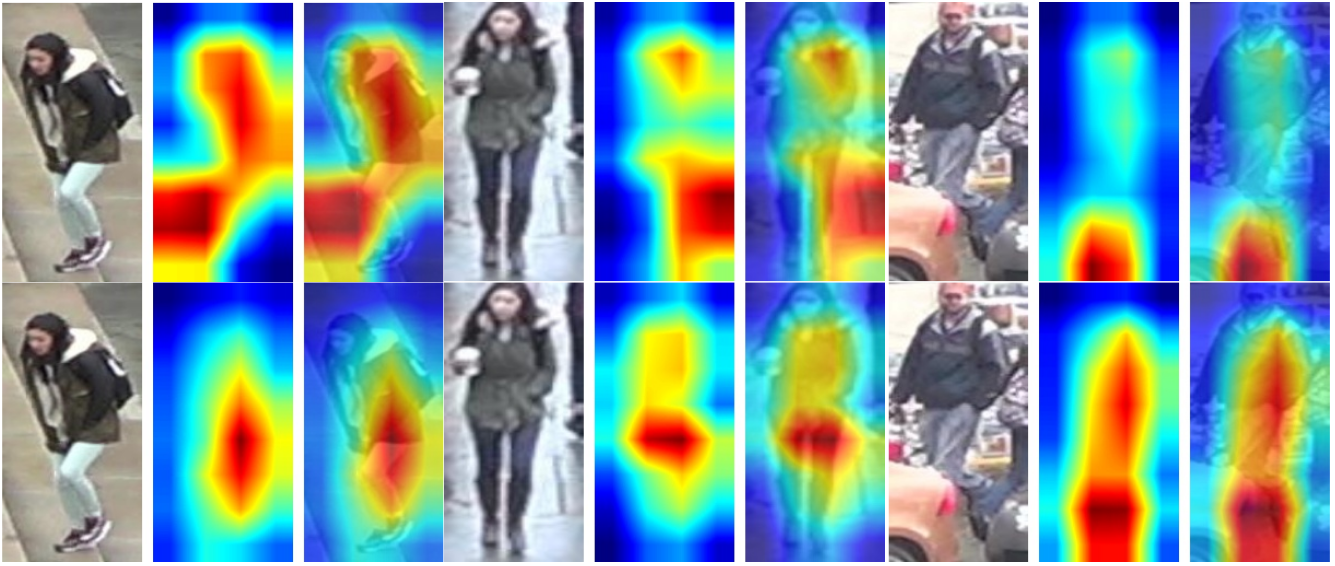}
                 \caption{Visualization of images from the target dataset. Top row is from the baseline SPCL and bottom is our proposed approach. The first image represent the sample, the middle is the activation map and the last image is the superposition of the sample with the activation map.}
                \label{fig:actmap}
            \end{figure*}

            Figure \ref{fig:actmap} displays examples of activation maps \cite{zagoruyko2016paying,aich2021spatio} for SPCL and our proposed method obtained from three bounding box images, featuring different individuals, in the target domain (ILIDS-VID dataset). The activation maps indicates the regions of interest of the backbone CNN when  extract feature representations.  The figure shows that our method provides a better localization of the person. Compared to the baseline SPCL, less background information is captured, allowing the model to focus on strong identity-based features.
            
\end{document}